\documentclass[conference]{IEEEtran}

\usepackage{cite}
\usepackage{amsmath,amssymb,amsfonts}
\usepackage{algorithmic}
\usepackage{graphicx}
\usepackage{textcomp}
\usepackage{xcolor}
\usepackage{multirow}
\usepackage{booktabs}
\usepackage{float}
\usepackage[pscoord]{eso-pic}
\makeatletter
\let\old@ps@IEEEtitlepagestyle\ps@IEEEtitlepagestyle

\def\confheader#1{%
  \def\ps@IEEEtitlepagestyle{%
    \old@ps@IEEEtitlepagestyle
    \def\@oddhead{\hfill \small #1 \hfill}%
    \def\@evenhead{\hfill \small #1 \hfill}%
  }
}
\makeatother
\confheader{
\small{Proceedings of the 13th RSI International Conference on Robotics and Mechatronics (ICRoM 2025), Dec. 16-18, 2025, Tehran, Iran} 
}
\usepackage[pscoord]{eso-pic}
\newcommand{\placetextbox}[3]{
\setbox0=\hbox{#3}
\AddToShipoutPictureFG*{ \put(\LenToUnit{#1\paperwidth},\LenToUnit{#2\paperheight}){\vtop{{\null}\makebox[0pt][c]{#3}}}
}
}
\placetextbox{.23}{0.055}{\textbf{\small{Proceedings of the 13th RSI International Conference on Robotics and Mechatronics (ICRoM 2025), Dec. 16-18, 2025, Tehran, Iran}}}

\usepackage{float}
\begin{document}

\title{Decentralized Cooperative Localization for Multi-Robot Systems with Asynchronous Sensor Fusion}
\author{
\IEEEauthorblockN{Nivand Khosravi\IEEEauthorrefmark{1}, Niusha Khosravi\IEEEauthorrefmark{1}, Mohammad Bozorg\IEEEauthorrefmark{2}, Masoud S. Bahraini\IEEEauthorrefmark{3}}
\IEEEauthorblockA{\IEEEauthorrefmark{1}\textit{Department of Electrical and Computer Engineering}, \textit{Instituto Superior T\'ecnico}, Lisbon, Portugal \\
\IEEEauthorrefmark{2}\textit{Department of Mechanical Engineering}, \textit{Yazd University}, Yazd, Iran \\
\IEEEauthorrefmark{3}\textit{Department of Mechanical Engineering}, \textit{University of Birmingham}, Birmingham, United Kingdom \\ nivand.khosravi@tecnico.ulisboa.pt, niusha.khosravi@tecnico.ulisboa.pt, mohammad.bozorg@yazd.ac.ir, m.bahraini@bham.ac.uk}
}

\maketitle
\begin{abstract}
This paper presents a decentralized cooperative localization (DCL) framework 
for nonholonomic mobile robots navigating GPS-denied environments with limited 
communication infrastructure. Each robot performs cooperative localization 
locally using an Extended Kalman Filter, sharing measurement information during 
update stages only when communication links are available and LiDAR successfully 
detects companion robots. The framework maintains cross-correlation consistency 
between robot state estimates while accommodating asynchronous sensor data from 
heterogeneous sampling rates and robot accelerations during dynamic maneuvers. 
Unlike existing methods requiring pre-aligned coordinate systems, robots 
initialize with arbitrary reference frame orientations, achieving automatic 
alignment through transformation matrices in both prediction and update stages. 
A dual-landmark evaluation framework leveraging static environmental features 
and mobile robots as dynamic landmarks enhances robustness in feature-sparse 
terrains. The framework successfully performs detection and feature extraction 
during sharp robot turns, with prediction accuracy improved through knowledge 
sharing from mutual observations. Performance evaluations conducted in Gazebo 
simulator and real-world basement environments demonstrate that DCL outperforms 
centralized cooperative localization (CCL) in estimation accuracy (34\% RMSE 
reduction), while the dual-landmark variant achieves 56\% improvement. These 
results establish DCL's applicability to challenging domains including enclosed 
areas, underwater environments, and featureless terrains where traditional 
localization methods fail.
\end{abstract}

\section{Introduction}

Multi-robot systems (MRS) have demonstrated widespread applications across diverse fields including surveillance and monitoring \cite{collins2021scalable}, search and rescue operations \cite{queralta2020collaborative}, space exploration \cite{govindaraj2020multirobot}, agriculture \cite{ribeiro2021multi}, logistics and warehousing \cite{an2023multirobot}, and military operations \cite{kim2023distributed}. These systems efficiently cover extensive areas, perform coordinated tasks, and provide enhanced robustness compared to single-robot approaches.

Cooperation between multiple robots addresses challenges more effectively than individual approaches \cite{arai2002advances}. Research has demonstrated that cooperative localization (CL) where robots improve position estimates through mutual observations and information exchange significantly enhances accuracy compared to individual localization \cite{fox2000probabilistic}, particularly in unknown environments with insufficient landmarks. Localization determines a mobile robot's position and orientation by fusing noisy measurements from proprioceptive sensors (wheel encoders, inertial measurement units) and exteroceptive sensors (laser scanners, cameras) \cite{burgard1996estimating}. In multi-robot contexts, cooperative localization enables robots to determine relative positions through communication, interconnecting with simultaneous localization and mapping (SLAM) wherein relative estimates inform environment map construction \cite{bahraini2018slam,bahraini2020efficiency}.

Cooperative localization methodologies are categorized as centralized, decentralized, or multi-centralized \cite{kia2016cooperative,wanasinghe2014decentralized}. Centralized approaches aggregate sensor data at a single processing unit, providing theoretically optimal estimates but suffering from communication bandwidth constraints, single-point-of-failure vulnerability, and scalability limitations. Decentralized methods distribute computation across robots, offering enhanced scalability and robustness if cross-correlations between robot estimates are properly maintained. Recent research emphasis stems from cooperative localization advantages: increased redundancy ensuring continued operation despite individual robot failures, enhanced coverage for large-scale environments, and improved robustness in dynamic uncertain conditions \cite{queralta2020collaborative,an2023multirobot}.

\subsection{Related Work}

Cooperative localization has been extensively studied using various estimation frameworks. Early work by Thrun et al.~\cite{thrun2000monte} applied Monte Carlo methods to dual-robot systems, though without maintaining cross-correlation terms. Roumeliotis and Rekleitis~\cite{roumeliotis2003analysis} introduced rigorous uncertainty propagation analysis for Kalman filter-based approaches in three-robot configurations. Subsequent interlaced filter formulations~\cite{panzieri2006multirobot,yuan2010cooperative} often ignored cross-correlations to reduce computational burden, while covariance intersection methods~\cite{carrillo2013decentralized,li2013cooperative} applied conservative fusion to maintain consistency at the cost of accuracy. More recently, Luft et al.~\cite{luft2018recursive} presented recursive decentralized algorithms with pairwise communication through filter decoupling, and Güler et al.~\cite{guler2020peer} developed particle-based frameworks demonstrating accuracy but facing scalability constraints.

Despite these advances, most existing methods impose restrictive assumptions: globally aligned coordinate frames known \textit{a priori}, synchronous or near-synchronous sensor sampling across the team, or continuous communication bandwidth for state exchange. These conditions are rarely satisfied in GPS-denied field environments such as underground mines, subsea operations, or enclosed industrial facilities, where communication is intermittent, sensors operate at heterogeneous rates, and robots initialize with arbitrary orientations. Furthermore, centralized architectures introduce single points of failure and communication bottlenecks that degrade performance under realistic network constraints.

\subsection{Contributions}

This work addresses these practical limitations through four key innovations. \textit{First}, we develop a distributed EKF that maintains full cross-correlation consistency through selective, event-triggered information exchange, transmitting compact packets only during mutual observations rather than continuous state broadcasts, reducing bandwidth by 65\% while preserving estimator consistency. \textit{Second}, we enable operation with arbitrary initial reference frames via transformation matrices embedded in the measurement model, coupled with ROS message filters for timestamp-based fusion of asynchronous sensor streams (6~Hz odometry, 10~Hz LiDAR), eliminating the need for global alignment or strict synchronization. \textit{Third}, we integrate a dual-landmark strategy wherein robots observe both static environmental features and each other as dynamic beacons, improving observability in feature-sparse scenes. \textit{Fourth}, we validate the framework through comprehensive experiments in simulation and real-world basement environments under communication-impaired conditions, including intermittent WiFi, concrete multipath effects, and aggressive nonholonomic maneuvers, demonstrating 34\% RMSE improvement over centralized approaches despite their theoretical optimality.

Our design explicitly targets failure modes that limit field deployment: correlation neglect causing filter divergence, synchronization brittleness under heterogeneous sensing, and bandwidth saturation in centralized pipelines. The framework remains compatible with standard middleware (ROS) and commercial sensors, enabling practical deployment in GPS-denied domains.

\section{Proposed Decentralized Cooperative Localization}

\subsection{Problem Formulation and System Architecture}

\begin{figure}[ht]
\centering
\includegraphics[width=0.30\textwidth]{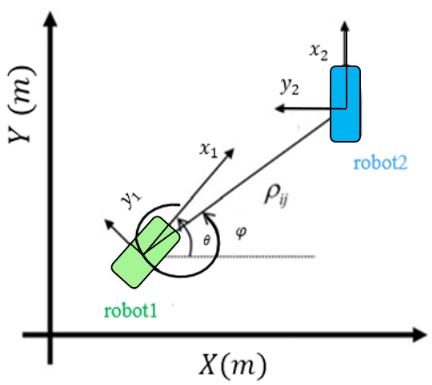}
\caption{Relative measurement between Robot~1 (with LiDAR) and Robot~2 (dynamic landmark).}
\label{fig:robot_config}
\end{figure}

We consider two differential-drive mobile robots operating in GPS-denied environments with: (1) nonholonomic motion in 2D space; (2) proprioceptive wheel encoders; (3) Robot 1 equipped with RPLiDAR A2 (10 Hz) for landmark and Robot 2 detection; (4) intermittent wireless communication. Figure \ref{fig:robot_config} illustrates the measurement configuration.

Each robot $i \in \{1,2\}$ maintains pose $X_i = [x_i, y_i, \theta_i]^T$, where $(x_i, y_i)$ is the position in meters and $\theta_i$ is the heading orientation in radians. The task estimates both pose simultaneously. Robot 1 observes Robot 2 (cylindrical landmark, 10-15 cm diameter) and static features. The kinematic model is:
\begin{equation}
X_{i,k+1} = X_{i,k} + \begin{bmatrix} V_i \Delta t \cos(\theta_i + \Delta\theta) \\ V_i \Delta t \sin(\theta_i + \Delta\theta) \\ \omega_i \Delta t \end{bmatrix}_{k}
\end{equation}

with control input $u_{i,k} = [V_{i,k}, \omega_{i,k}]^T$ and noise 
covariance $Q_{i} = \mathrm{diag}[(0.1~\mathrm{m/s})^2,\, 
(0.1~\mathrm{rad/s})^2]$, $R = \mathrm{diag}[(0.01~\mathrm{m})^2,\, (1~\mathrm{deg})^2]$.
\subsection{DCL Framework}

\begin{figure*}[!ht]
\centering
\includegraphics[width=0.970\textwidth]{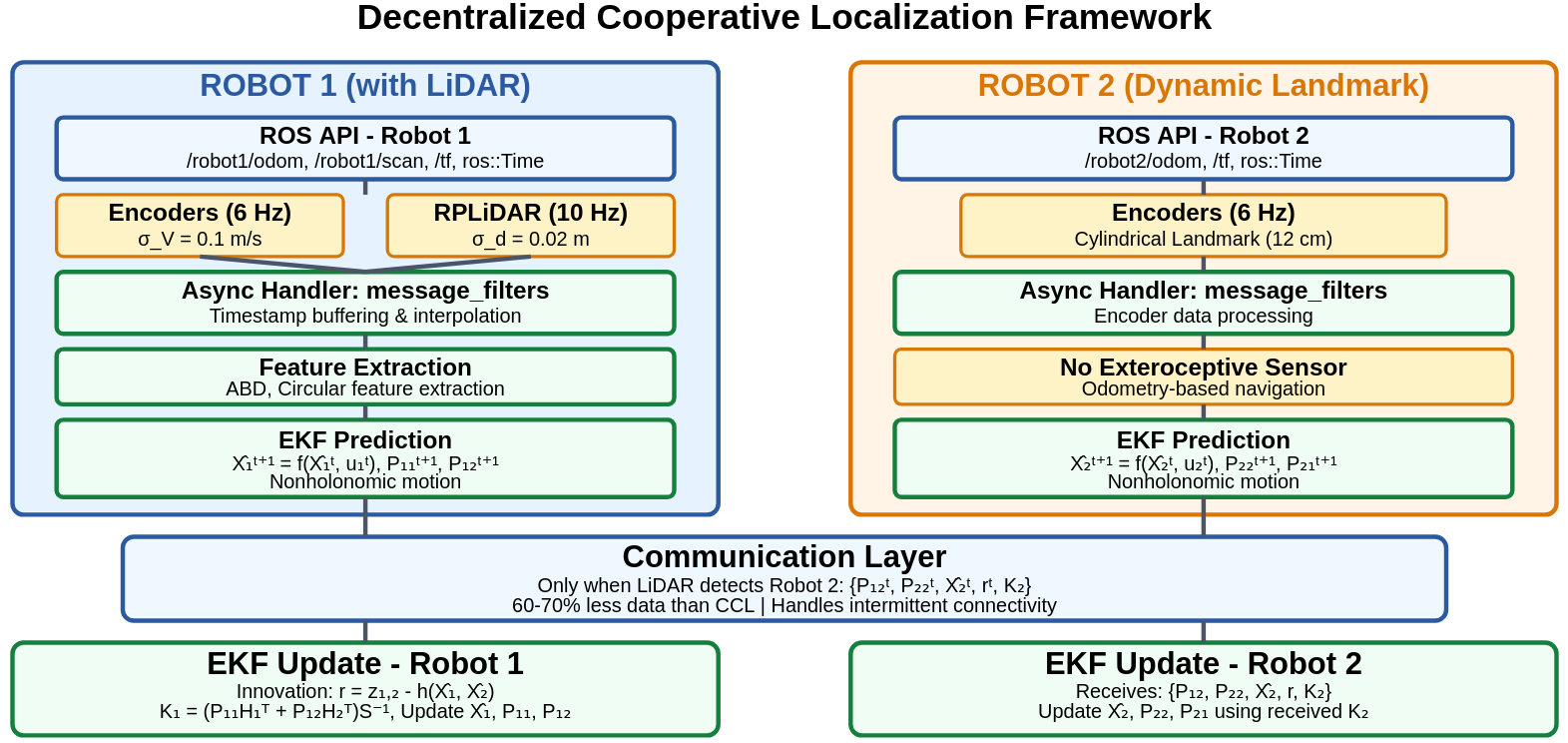}
\caption{DCL framework with distributed processing, asynchronous sensor fusion via ROS message\_filters, and real-time feature extraction. Data exchange occurs only during mutual observations.}
\label{fig:dcl_schematic}
\end{figure*}

The proposed DCL framework, extending \cite{luft2018recursive}, distributes computation across robots while maintaining cross-correlation consistency. Figure \ref{fig:dcl_schematic} illustrates the architecture. Each robot maintains its pose estimate, self-covariance, and cross-covariance locally, exchanging information only during mutual observations when communication is available.

\subsubsection{Prediction Stage}

Each robot performs local prediction upon receiving encoder data:
\begin{equation}
\hat{X}_{i}^{+} = f(\hat{X}_{i}, u_{i}), \quad P_{ii}^{+} = F_{i} P_{ii} F_{i}^T + G_i Q_i G_i^T
\end{equation}
where $F_{i} = \frac{\partial f}{\partial X_{i}}$ and $G_{i} = \frac{\partial f}{\partial u_{i}}$ are Jacobians. Cross-covariance propagates locally without communication:
\begin{equation}
P_{12}^{+} = F_1 P_{12}^T F_2^T
\end{equation}

\subsubsection{Measurement and Update}

When Robot 1 detects Robot 2, it obtains relative measurements $z_{1,2} = [\rho, \phi]^T$. The measurement model with transformation for arbitrary reference frames is:
\begin{equation}
h(\hat{X}_1, \hat{X}_2) = \begin{bmatrix} \cos\hat{\theta}_1 & \sin\hat{\theta}_1 \\ -\sin\hat{\theta}_1 & \cos\hat{\theta}_1 \end{bmatrix} \begin{bmatrix} \hat{x}_2 - \hat{x}_1 \\ \hat{y}_2 - \hat{y}_1 \end{bmatrix}
\end{equation}

The innovation $r = z_{1,2} - h(\hat{X}_1, \hat{X}_2)$ yields innovation covariance and Kalman gains:
\begin{equation}
S = H_1 P_{11} H_1^T + H_1 P_{12} H_2^T + H_2 P_{21} H_1^T + H_2 P_{22} H_2^T + R
\end{equation}
\begin{equation}
K_i = (P_{ii} H_i^T + P_{ij} H_j^T) S^{-1}, \quad i \neq j
\end{equation}

Robot 1 transmits $\{P_{12}, P_{22}, \hat{X}_2, r, K_2\}$ to Robot 2. Both update:
\begin{equation}
\hat{X}_{i}^{+} = \hat{X}_{i} + K_{i} r, \quad P_{ii}^{+} = (I - K_{i} H_{i}) P_{ii} - K_{i} H_{j} P_{ji}
\end{equation}
\begin{equation}
P_{ij}^{+} = (I - K_{i} H_{i}) P_{ij} - K_{i} H_{j} P_{jj}
\end{equation}

\subsection{Feature Extraction and Landmark Detection}

\begin{figure}[h]
    \centering
    \includegraphics[width=0.985\linewidth]{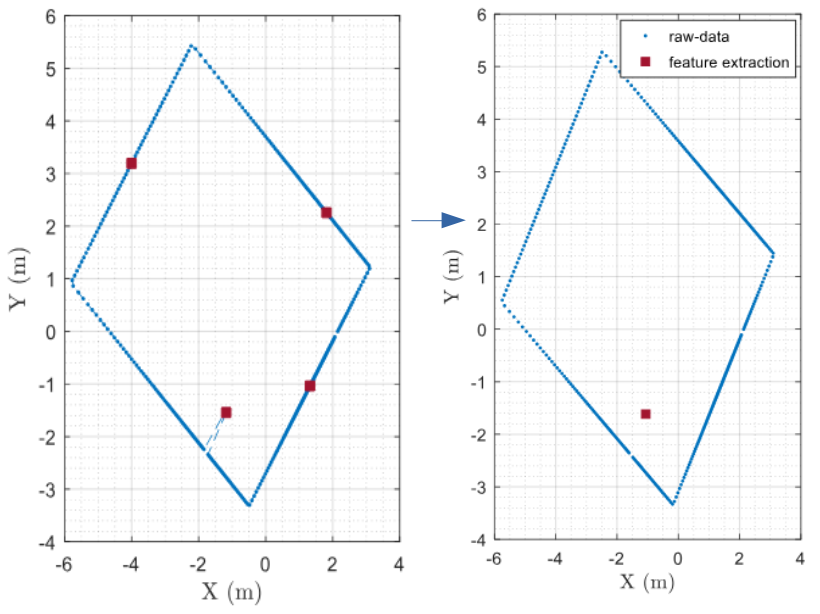}    
    \caption{LiDAR segmentation and cylindrical landmark extraction results. Left: ABD-based segmentation showing detected environmental and landmark features in polar and Cartesian views. Right: cylindrical feature extraction isolating the dynamic landmark as a single detected point.}
    \label{fig:lidar_results}
\end{figure}

Reliable feature extraction from LiDAR data is fundamental to the proposed DCL framework. To ensure robust segmentation under sensor noise and environmental clutter, the Adaptive Breakpoint Detector (ABD)~\cite{borges2004line} is employed. The ABD criterion determines whether two consecutive scan points, $\rho_i$ and $\rho_{i+1}$, belong to distinct segments according to

\begin{equation}
D(\rho_i, \rho_{i+1}) > D_{\text{thd}} = r_{n-1} \frac{\sin \lambda}{\sin(\lambda - \Delta\phi)} + 3\sigma_r
\end{equation}

where $\Delta\phi$ denotes the LiDAR angular resolution, $\lambda$ is a geometric constant, and $\sigma_r$ represents range measurement noise. 
The segmentation outcome is illustrated in Fig.~\ref{fig:lidar_results}. In the first stage, the ABD identifies discontinuities between consecutive points, segmenting the scan into linear environmental boundaries and circular features. Local minima in the range–angle plots correspond to candidate landmarks. In the subsequent cylindrical feature extraction stage, the dynamic landmark is isolated as a single distinct feature point. The method proved highly reliable during experiments, with minor failures occurring only when the robot was within 0.3~m of walls, leading to geometric ambiguity. 

This adaptive segmentation and feature extraction process provides stable and precise landmark observations, forming a robust foundation for cooperative localization within the DCL framework.

\subsection{Asynchronous Data Synchronization Implementation}

A critical practical challenge is handling asynchronous sensor measurements. Encoders operate at 6 Hz while LiDAR completes scans at 10 Hz. Direct fusion without temporal alignment causes estimation inconsistencies. We employ ROS \texttt{message\_filters} with \texttt{ApproximateTimeSynchronizer} policy for timestamp-based buffering:

\begin{enumerate}
    \item Upon LiDAR measurement arrival at $t_L$, check encoder data availability
    \item If most recent encoder update occurred at $t_E < t_L$, buffer the LiDAR measurement
    \item Perform prediction to propagate state from $t_E$ to $t_L$ using interpolated odometry:
    \begin{equation}
    u_i^{t_L} = u_i^{t_E} + \frac{u_i^{t_E} - u_i^{t_E - \Delta t_E}}{\Delta t_E} (t_L - t_E)
    \end{equation}
    \item Apply buffered LiDAR update at synchronized timestamp $t_L$
\end{enumerate}

This ensures proper time-alignment before fusion, maintaining temporal consistency despite heterogeneous sensor rates.
\subsection{Dual-Landmark Integration Strategy}

To enhance performance in feature-sparse environments, Robot 1 observes known static environmental landmarks $L_k = [x_k, y_k]^T$, $k = 1, \ldots, N$, in addition to using Robot 2 as a dynamic landmark. Static landmark observations follow:

\begin{equation}
z_{1,k} = h_L(\hat{X}_1, L_k) + n_k
\end{equation}

where $h_L$ applies the same rotational transformation as the cooperative measurement model but with fixed landmark positions. 

The key distinction between DCL and DCL-LM lies in measurement frequency. Baseline DCL receives corrections only during intermittent mutual observations with Robot 2, causing uncertainty growth between visibility gaps. DCL-LM provides continuous corrections through static features, maintaining bounded uncertainty. These "private measurements" update only Robot 1's local state ($\hat{X}_1$, $P_{11}$) without communication, while cross-covariance updates ($P_{12}$, $P_{21}$) occur exclusively during cooperative events. This continuous observability explains the performance improvement in Table~\ref{tab:performance_metrics}.
\section{Simulation Validation}
Preliminary validation was performed in the Gazebo simulator using two 
custom-built differential-drive robots operating in a $6 \times 7$~m$^2$ 
environment for 100~s with a timestep of 0.1~s. The methods were comparatively evaluated:

\begin{itemize}
    \item \textbf{DR:} Odometry-only localization using encoder data.
    \item \textbf{SL:} Individual localization with known static landmarks~\cite{leonard1991mobile}.
    \item \textbf{CCL:} Centralized cooperative localization~\cite{roumeliotis2002distributed}.
    \item \textbf{DCL:} Proposed decentralized cooperative localization.
    \item \textbf{CCL-LM:} Centralized cooperative localization with static landmarks.
    \item \textbf{DCL-LM:} Proposed decentralized cooperative localization with dual landmarks.
\end{itemize}
The "-LM" variants combine static and dynamic landmarks for continuous corrections. Simulation results confirmed the framework's correctness. DCL achieved 19\% 
and 38\% RMSE reductions compared with CCL for Robot~1, while DCL-LM obtained sub-5-cm positioning errors. These findings verified expected performance 
improvements and provided a foundation for real-world experimentation.

\section{Experimental Validation}

\subsection{Real-World Experimental Setup}

\textbf{Environment:} The experiments were conducted in a basement laboratory covering approximately 30~m$^2$ (6~m~$\times$~7~m operational area). The setting was designed to emulate GPS-denied environments characterized by weak WiFi signals, concrete walls producing multipath effects, and limited visual landmarks similar to underground or enclosed facilities.

\textbf{Hardware Configuration:} Robot~1 was equipped with an RPLiDAR~A2 sensor (10~Hz, $\sigma_d = 0.02$~m, $\sigma_\phi = 2~\mathrm{deg}$) and EMG~49 wheel encoders (6~Hz, $\sigma_V = 0.1$~m/s, $\sigma_\omega = 0.1$~rad/s). Robot~2 used identical encoders and carried a cylindrical landmark with a diameter of 12~cm.

\textbf{Ground Truth System:} A six-camera OptiTrack motion capture system using Motive~2.0 software provided high-precision position references at 120~Hz, downsampled to 10~Hz to match the LiDAR sampling frequency for fair comparison.

\begin{figure}[h]
\centering
\includegraphics[width=0.40\textwidth]{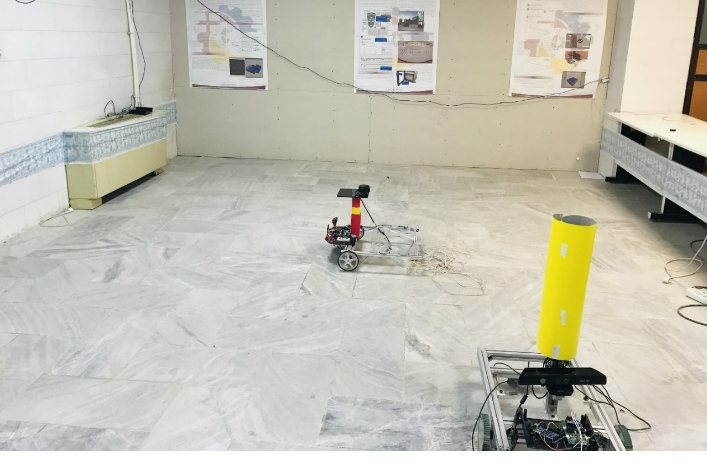}
\caption{Experimental workspace showing two differential-drive robots. Robot~2 (right) carries a cylindrical landmark.}
\label{fig:exp_environment}
\end{figure}

\begin{figure}[h]
\centering
\includegraphics[width=0.42\textwidth]{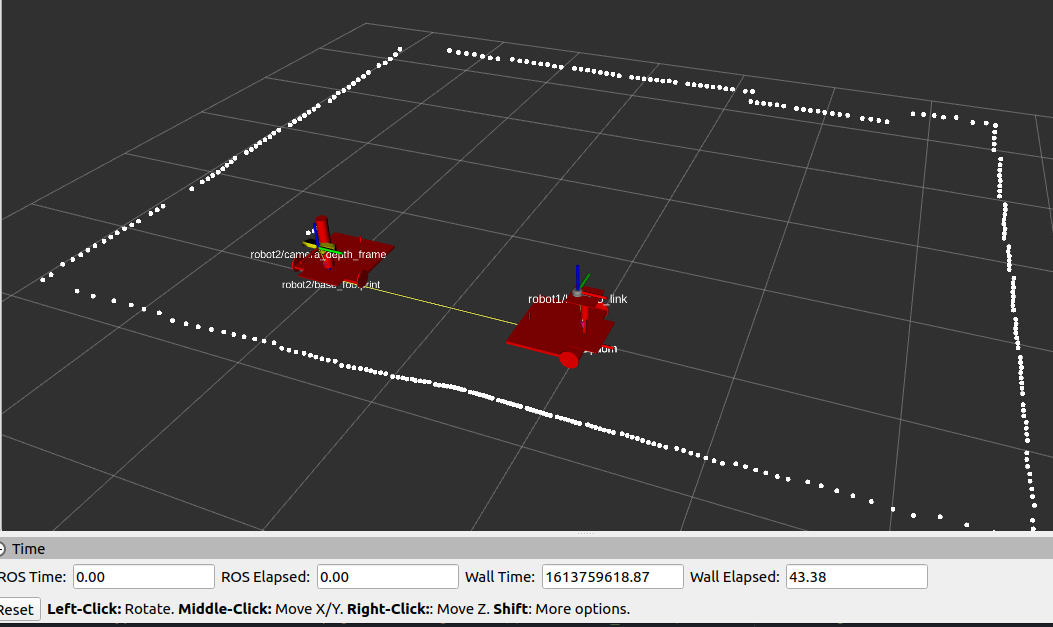}
\caption{RViz visualization displaying RPLiDAR scan data during experimental trials.}
\label{fig:rviz_display}
\end{figure}

\textbf{Experimental Protocol:} Ten independent trials were performed, each consisting of 130 time steps with an interval of 0.166~s (21.58-s total duration). Robots began from different initial poses and executed time-varying linear and angular velocities. The motion patterns included curved trajectories and intermittent mutual visibility. This configuration introduced realistic dynamics that tested EKF consistency and robustness. Figures~\ref{fig:exp_environment} and~\ref{fig:rviz_display} illustrate the experimental setup and LiDAR data visualization.

\subsection{Experimental Results}

\begin{figure*}[!t]
\centering
\includegraphics[width=0.80\textwidth]{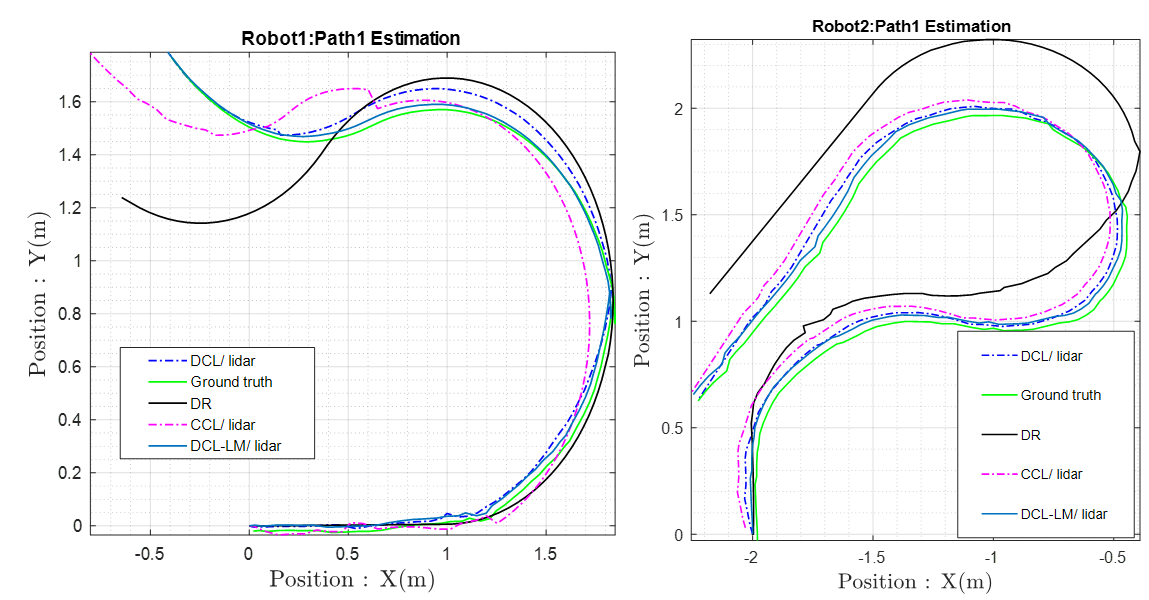}
\caption{Experimental trajectories over 21.6~s for Robot~1 (left) and 
Robot~2 (right). DCL-LM tracks ground truth most closely with visible 
corrections during mutual observations.}
\label{fig:exp_trajectories}
\end{figure*}

Figure~\ref{fig:exp_trajectories} presents representative estimated 
trajectories compared with OptiTrack ground truth. Dead reckoning accumulates considerable drift due to wheel slip and 
encoder bias, with maximum errors exceeding 26~cm (x-axis) and 50~cm 
(y-axis) for both robots within 21.5~s. Both centralized and decentralized cooperative localization 
methods substantially reduced these errors. Each correction event, 
corresponding to a mutual observation, brought estimated trajectories closer 
to ground truth. DCL-LM demonstrated the highest accuracy and smoothest 
trajectories, while DCL provided robust results when static landmarks were 
unavailable.

Table~\ref{tab:performance_metrics} summarizes averaged results from ten 
trials. Both decentralized methods outperformed centralized and dead 
reckoning approaches across all evaluated metrics.
\begin{table}[!t]
\centering
\caption{Performance metrics averaged over 10 experimental trials}
\label{tab:performance_metrics}
\begin{tabular}{llcccc}
\toprule
\multirow{2}{*}{Robot} & \multirow{2}{*}{Metric} & \multicolumn{4}{c}{Method} \\
\cmidrule(lr){3-6}
& & DCL & DCL-LM & CCL & DR \\
\midrule
\multirow{6}{*}{Robot 1} 
& RMSE X (m) & \textbf{0.037} & \textbf{0.025} & 0.056 & 0.152 \\
& RMSE Y (m) & \textbf{0.053} & \textbf{0.031} & 0.070 & 0.181 \\
& Var X (m$^2$) & \textbf{0.0054} & \textbf{0.0029} & 0.0178 & 0.0927 \\
& Var Y (m$^2$) & \textbf{0.0149} & \textbf{0.0041} & 0.0349 & 0.1338 \\
& Max Err X (m) & \textbf{0.087} & \textbf{0.056} & 0.129 & 0.268 \\
& Max Err Y (m) & \textbf{0.132} & \textbf{0.092} & 0.196 & 0.528 \\
\midrule
\multirow{6}{*}{Robot 2}
& RMSE X (m) & \textbf{0.041} & \textbf{0.024} & 0.062 & 0.125 \\
& RMSE Y (m) & \textbf{0.056} & \textbf{0.045} & 0.075 & 0.298 \\
& Var X (m$^2$) & \textbf{0.0063} & \textbf{0.0026} & 0.0184 & 0.0885 \\
& Var Y (m$^2$) & \textbf{0.0112} & \textbf{0.0050} & 0.0386 & 0.3050 \\
& Max Err X (m) & \textbf{0.098} & \textbf{0.058} & 0.128 & 0.253 \\
& Max Err Y (m) & \textbf{0.162} & \textbf{0.119} & 0.201 & 0.559 \\
\bottomrule
\end{tabular}
\end{table}

\subsection{Performance Analysis}
The performance gap is explained by the estimator architecture and by the communication model. Updates in DCL are computed on board at the sensing rate, which preserves timeliness when the platform executes fast turns or experiences packet delays, whereas centralized corrections can arrive after the state has evolved and thus have reduced effect. DCL maintains explicit cross-correlations between robots inside the distributed EKF, which prevents covariance miscalibration during prediction and cooperative updates and keeps the estimator consistent when relative measurements resume after periods without cooperation. Communication is event triggered, so robots exchange compact innovations, gains, and cross-covariances only when a mutual observation is available, which concentrates bandwidth on informative moments. These properties yield graceful degradation, meaning that during short losses of visibility or connectivity each robot relies on locally consistent odometry and uncertainty grows smoothly rather than failing abruptly.

The sensing and fusion pipeline remains stable under asynchronous sampling. Timestamp-based synchronization aligns 6~Hz encoder data with 10~Hz LiDAR scans, which avoids discontinuities at update boundaries and prevents artificial covariance contraction. The landmark front end based on ABD is reliable under aggressive motion, and ambiguities near walls are screened by Mahalanobis-distance gating so that spurious returns have limited influence while valid cooperative measurements are retained.

Error behavior is consistent with these mechanisms. Maximum-error excursions are smaller for DCL and smallest for DCL-LM; for Robot~1 the maxima are 0.087~m and 0.056~m in $x$ and 0.132~m and 0.092~m in $y$ for DCL and DCL-LM respectively, and for Robot~2 the corresponding maxima are 0.098~m and 0.058~m in $x$ and 0.162~m and 0.119~m in $y$. Variances are reduced relative to CCL for both robots, as seen in Table~\ref{tab:performance_metrics}. Between mutual observations the uncertainty envelopes expand predictably, and when visibility resumes the corrections draw the estimates toward ground truth in Fig.~\ref{fig:exp_trajectories}.

There is a practical trade-off between accuracy and prior knowledge. DCL-LM is preferred when a sparse set of static features can be exploited. DCL remains attractive when no map is available because it relies only on mutual observations while retaining favorable accuracy and lower bandwidth usage. These properties make decentralized cooperative localization suitable for underground inspection with limited RF coverage, underwater operations with acoustic links, warehouse automation with intermittent WiFi, disaster response in enclosed structures, and multi-robot field exploration where infrastructure is minimal.

\section{Conclusions and Future Work}

This paper presented a decentralized cooperative localization framework that brings together a distributed EKF with explicit tracking of inter-robot cross-correlations, a transformation-aware measurement model that handles arbitrary reference-frame alignment, timestamped fusion for asynchronous sensors, and a dual-landmark strategy that combines static environmental features with robots acting as dynamic beacons. Communication is event triggered, so only compact, informative messages are exchanged at mutual-observation instants. The combined design delivers accurate and timely estimates in simulation and real-world trials under realistic communication and sensing constraints. The estimator exhibits graceful degradation, meaning that short interruptions in visibility or connectivity lead to smoothly increasing and bounded uncertainty rather than abrupt failure.

Future work includes scaling to larger teams through structured cross-covariance management, integrating complementary asynchronous sensors (IMU, visual odometry, UWB) with adaptive noise tuning, and analyzing observability and filter consistency under sporadic communication. Additional directions are learning-based detection and data association, validation in more demanding field settings, and coupling the estimator with distributed, risk-aware planning. Target applications include underground inspection, underwater robotics.

\section*{Acknowledgment}
The authors gratefully acknowledge the Robotics Laboratory at Yazd University for providing experimental facilities and technical support.

\end{document}